\pdfoutput=1

\documentclass[runningheads]{llncs}

\usepackage{eccv}
\usepackage{eccvabbrv}

\usepackage{graphicx}
\usepackage{booktabs}
\usepackage{multirow}
\usepackage{arydshln}
\usepackage{tabularx}
\usepackage{array}
\usepackage[table]{xcolor}
\usepackage{makecell}
\usepackage[accsupp]{axessibility}

\usepackage{hyperref}

\begin{document}

\title{GSMap: 2D Gaussians for Online HD Mapping}

\titlerunning{2D Gaussians for Online HD Mapping}

\author{Zhenxuan Zeng\inst{1,2} \and
Lingxuan Wang\inst{2} \and
Sheng Yang\inst{2}\thanks{Corresponding author.} \and
Yanan He\inst{1} \and \\
Mingxia Chen\inst{2} \and
Wei Suo\inst{1} \and
Peng Wang\inst{1}
}

\authorrunning{Z.~Zeng et al.}

\institute{
School of Computer Science, Northwestern Polytechnical University, China \and
Unmanned Vehicle Dept, Cainiao Inc., Alibaba Group, China\\
\email{zengzhenxuan@mail.nwpu.edu.cn, shengyang@cainiao.com}
}

\maketitle

\begin{abstract}
  Accurate High-Definition (HD) map construction is critical for autonomous driving, yet existing methods face a fundamental trade-off: vectorization-based approaches preserve topology but struggle with geometric fidelity, while rasterization-based approaches enable precise geometric supervision but produce unstructured outputs. To bridge this gap, we propose GSMap, a novel framework that unifies both paradigms via a learnable 2D Gaussian representation. Each map element is modeled as an ordered sequence of 2D Gaussians, whose centers correspond to the vertices of the vectorized polyline/polygon. This formulation enables simultaneous optimization through: (1) Differentiable rasterization that enforces pixel-level geometric constraints, and (2) Topology-aware vectorization that maintains structural regularity. Experiments on both nuScenes and Argoverse2 demonstrate that our Gaussian-based representation effectively unifies geometric and topological learning, achieving significant performance improvements and demonstrating strong compatibility with existing HD mapping architectures. Code will be available at \href{https://github.com/peakpang/GSMap}{https://github.com/peakpang/GSMap}
\end{abstract}

\section{Introduction}
\label{sec:intro}
\begin{figure*}[ht]
    \centering
    \includegraphics [width=1\linewidth]{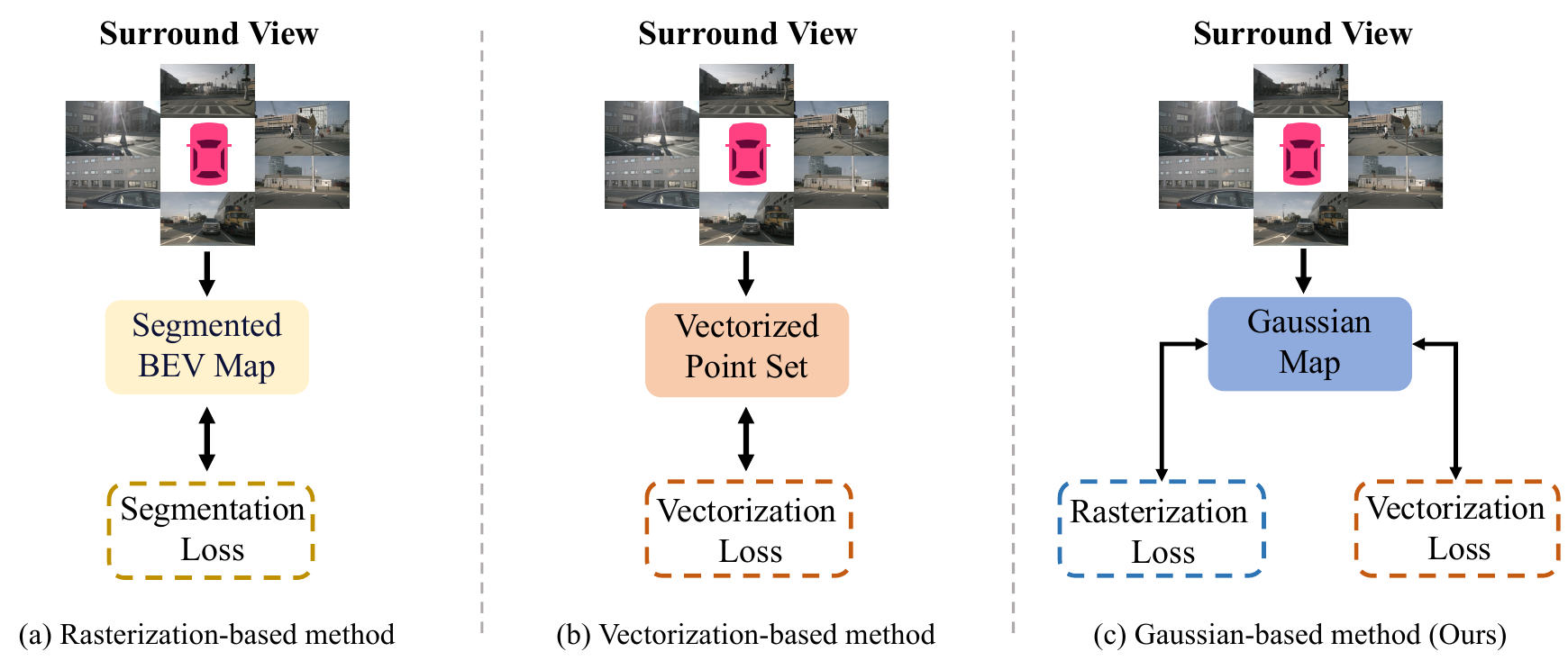}
    \caption{
    Comparison of online HD map construction paradigms. 
    (a) \textbf{Rasterization-based methods} formulate map construction as BEV segmentation, providing dense pixel-level supervision but yielding unstructured outputs. 
    (b) \textbf{Vectorization-based methods} directly predict ordered point sets, naturally preserving topology but lacking dense geometric supervision. 
    (c) \textbf{Our GSMap framework} unifies both paradigms by representing each map element as a learnable set of 2D Gaussians. This design enables simultaneous optimization via differentiable rasterization and topology-aware vectorization, bridging the gap between geometric fidelity and structural regularity.
    }
    \label{fig:motivation}
\end{figure*}
High-Definition (HD) maps serve as the cornerstone of autonomous driving systems, providing centimeter-accurate representations of road geometry and topology essential for safe navigation~\cite{prakash2021multi, chen2024end, da2022path, gao2020vectornet, he2024egovm, hu2023planning, jiang2022perceive}. While traditional map construction~\cite{loam,lego-loam,lio-sam} relies on costly offline pipelines, recent advances have reformulated it as an online Bird's-Eye-View (BEV) perception~\cite{roddick2020predicting, liang2022bevfusion, li2024bevformer, Bevsegformer} problem. 

Building on these advances, existing learning-based approaches for online HD mapping mainly fall into two distinct paradigms (Fig.~\ref{fig:motivation}): (1) \textbf{Rasterization-based approaches}~\cite{Hdmapnet,Superfusion,Mapprior,xiong2023neural} formulate map construction as BEV segmentation, leveraging dense pixel-wise supervision to capture geometric details. Although effective in learning spatial distributions, these methods produce unstructured outputs that require heuristic post-processing to extract usable vector elements (\eg, pedestrian crossing, lane divider and road boundary), limiting the direct applicability to downstream modules. \textbf{(2) Vectorization-based methods}~\cite{Vectormapnet, MapTR, Mapprior, Pivotnet, Streammapnet, Maptrv2, Mgmap, Himap, Mgmapnet, InteractionMap, polydiffuse, rtmap}, in contrast, directly predict ordered point sequences that naturally preserve topology and instance structure. However, they rely on sparse point-level regression losses that lack continuous geometric supervision.

Recent efforts attempt to bridge this gap by incorporating differentiable rasterization into vectorized frameworks.  For example, MapVR~\cite{MapVR} introduces rasterization-based geometric supervision to assist vectorization-based learning, improving accuracy without increasing the inference cost. However, the underlying representation in these methods remains a discrete point sequence, while the rasterization serves only as an auxiliary training signal. Consequently, geometry and topology are still modeled in separate spaces, preventing coherent joint optimization.  

To bridge this division, we propose GSMap, a unified framework that models HD map elements as ordered sequences of learnable 2D Gaussians. Our key insight is that Gaussian primitives naturally unify rasterized and vectorized supervision within a single, differentiable representation. Specifically: (1) \textbf{Differentiable rasterization} -- Each 2D Gaussian is rendered into a continuous BEV mask, providing dense, pixel-level geometric supervision. (2) \textbf{Topology-aware vectorization} -- the Gaussian centers directly correspond to the vertices of polylines or polygons, inherently maintaining order and structural regularity. Based on this shared representation, GSMap enables coherent joint optimization of geometric fidelity and topological correctness. Experiments on nuScenes and Argoverse2 demonstrate that GSMap consistently outperforms previous hybrid approaches, achieving stronger geometric accuracy while preserving topology. In addition to quantitative improvements, our framework introduces a unified representation paradigm for online HD map construction, enabling rasterization and vectorization to be learned coherently within a shared representation space. Our main contributions are summarized as follows:
\begin{itemize}
    \item We propose GSMap, a unified Gaussian-based representation that models each HD map element as an ordered sequence of learnable 2D Gaussians. 

    \item Building upon this representation, we introduce a joint optimization scheme that unifies rasterization-based geometric supervision and vectorization-based topological learning through a shared Gaussian representation.

    \item Extensive experiments on nuScenes and Argoverse2 demonstrate that GSMap achieves state-of-the-art performance among hybrid approaches.
\end{itemize}

\section{Related Work}
\label{sec:Related Work}

\subsection{Online HD Map Construction}

With the rapid development of autonomous driving, the real-time construction of high-definition (HD) maps has become increasingly critical. By enabling real-time perception of the surrounding environment and providing structured road geometry and semantics, HD maps significantly enhance driving safety. We refer readers to a comprehensive overview\cite{lyu2025online} of recent online HD mapping methods. From the perspective of supervision paradigms, existing methods can be broadly categorized into three classes: rasterization-based, vectorization-based, and hybrid representations.

\noindent\textbf{Rasterization-based methods.} Early works~\cite{Hdmapnet, Mapprior, xiong2023neural} formulate HD map construction as a semantic segmentation task in the bird’s-eye-view (BEV) space.  
This paradigm leverages dense pixel-level supervision to capture detailed geometry.  
For example, HDMapNet~\cite{Hdmapnet} transforms multi-view features into BEV space and applies multi-task segmentation heads to predict rasterized semantic maps. Neural Map Prior~\cite{xiong2023neural} maintains a global neural map representation in sparse tiles and uses cross-attention with GRU-based fusion to incorporate historical observations for enhanced online rasterized map inference. MapPrior~\cite{Mapprior} combines discriminative perception with a learned VQGAN-based generative prior to sample diverse, realistic, and uncertainty-aware rasterized layouts.
Although effective at recovering local geometric structures, these methods lack explicit modeling of map topology and connectivity, which limits their ability to represent structured road elements.

\noindent\textbf{Vectorization-based methods.} Recent approaches~\cite{Vectormapnet, MapTR, Mapprior, Pivotnet, Streammapnet, Maptrv2, Mgmap, Himap, Mgmapnet, InteractionMap, polydiffuse, rtmap} represent HD maps as sequences of ordered points, inspired by the DETR paradigm for structured prediction.  
This vectorized representation offers higher structural expressiveness but introduces optimization challenges such as uniform sampling bias and unstable point ordering.  
MapTR~\cite{MapTR} mitigates these issues with a permutation-equivalent modeling strategy, improving training stability and instance consistency.
VectorMapNet~\cite{Vectormapnet} employs a DETR-like~\cite{DETR} detector with learnable element queries and bipartite matching to first predict map element keypoints, then generates ordered polylines via an autoregressive transformer in a fully end-to-end manner.
Subsequent works like MapTRv2~\cite{Maptrv2} further enhance BEV representation learning through auxiliary segmentation supervision.  
However, the final predictions remain fully vectorization-based, with the rasterization signal merely acting as a weak auxiliary constraint during training.

\noindent\textbf{Hybrid representations.}  
To bridge the gap between the two paradigms, recent studies explore joint learning from both vectorized and rasterized representations.  
MapVR~\cite{MapVR} takes an important step in this direction by introducing differentiable rasterization to provide rendering-based supervision for vector predictions.  
This design allows mutual guidance between structural and pixel-level cues, yielding notable performance gains under both Chamfer-based and IoU-based metrics.  
However,, in MapVR, the rasterization branch is still auxiliary, and the underlying representation remains discrete. Our GSMap further develops this hybrid approach by introducing a unified Gaussian representation that intrinsically connects vectorized geometry and rasterized appearance within a single differentiable formulation, enabling truly joint optimization of both representations.

\begin{figure*}[t!]
    \centering
    \includegraphics [width=1\linewidth]{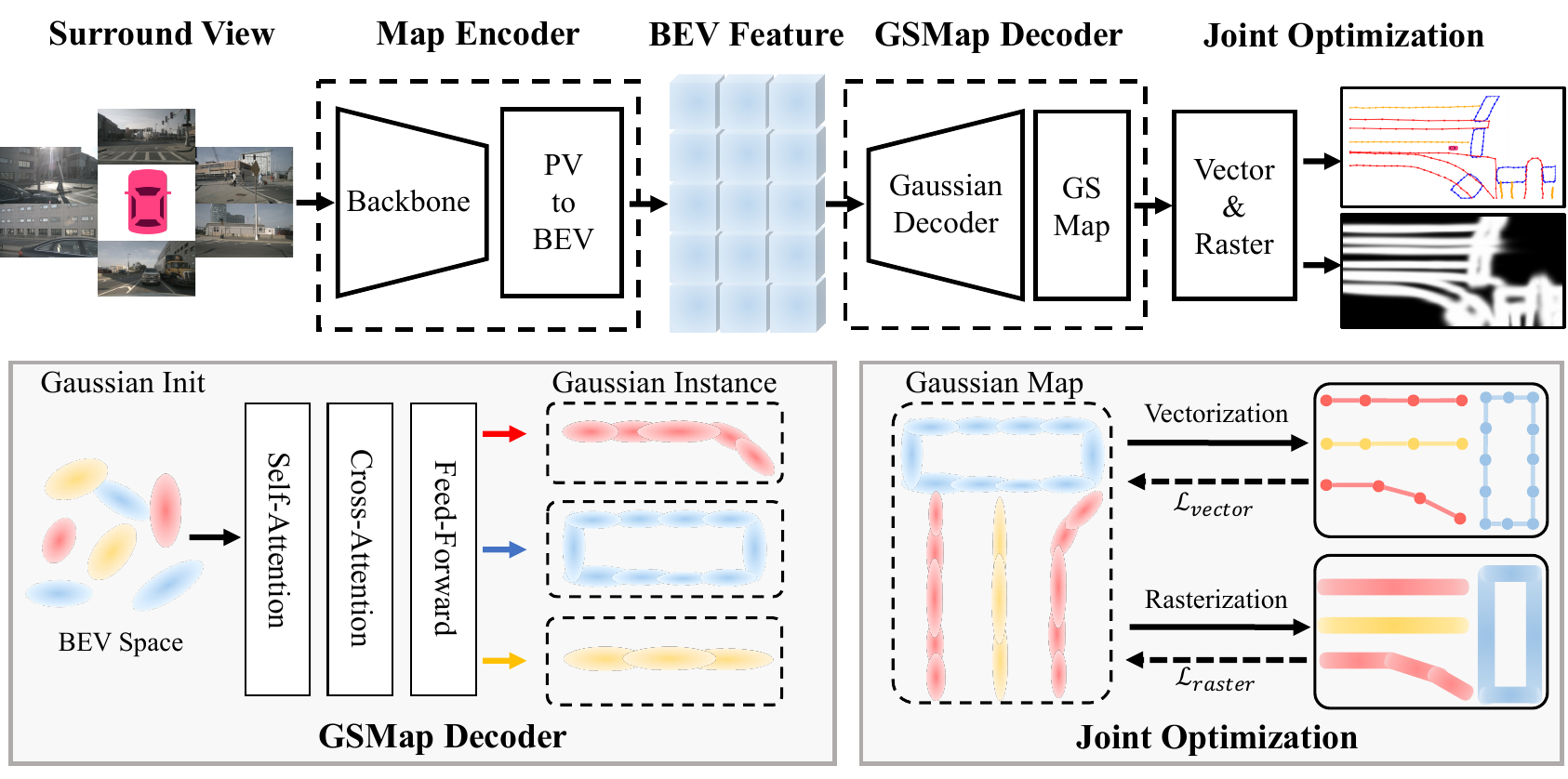}
    \caption{\textbf{Overview of GSMap.} First, the surrounding RGB images are fed into the Map Encoder to transform them into a unified BEV representation. Subsequently, GSMap initializes a set of instance queries composed of 2D Gaussians in the BEV space, which are refined by the GSMap Decoder to produce a unified Gaussian map. Each instance-level Gaussian sequence is (i) rasterized to an instance BEV mask via differentiable Gaussian rendering and (ii) converted to a vectorized polyline by connecting the ordered Gaussian centers. }
    \label{fig:pipeline}
\end{figure*}

\subsection{Gaussian Representation} 
Gaussian primitives have recently gained significant attention as an effective representation in computer vision and graphics. In 3D vision, Gaussian representations~\cite{kerbl20233d,huang20242d} have surpassed NeRF~\cite{mildenhall2021nerf} in novel view synthesis (NVS) tasks, demonstrating their efficiency and scalability. In autonomous driving, Gaussian-based representations have also been widely studied, with works such as GaussianFormer~\cite{huang2024gaussianformer} and GaussianBEV~\cite{chabot2025gaussianbev} showcasing the flexibility of Gaussians as continuous and differentiable primitives capable of bridging discrete and continuous representations.

However, to the best of our knowledge, Gaussian parameterizations have not yet been sufficiently explored in the context of HD map construction. Our work fills this gap by proposing a tailored 2D Gaussian representation for BEV map elements, thereby unifying vectorized and rasterized supervision within a single consistent framework.

\section{Method}
In this work, we propose GSMap, a unified framework for online high-definition (HD) map construction that bridges rasterization-based and vectorization-based paradigms through a learnable 2D Gaussian representation, and can be seamlessly integrated with existing HD map learning architectures.

\subsection{Overview}
Given multi-view surround images as input, GSMap constructs HD maps in an end-to-end manner by representing each map element as an ordered sequence of learnable 2D Gaussian primitives in the BEV space, as illustrated in Fig.~\ref{fig:pipeline}. This unified representation allows geometric learning in both rasterization and vectorization within a single differentiable formulation. 

Specifically, we first define the Gaussian-based representation (Sec.\ref{sec:Gaussian-based Representation}), where each Gaussian encodes spatial position, scale, and orientation.
Building upon this representation, we introduce two complementary supervision pathways (Sec.\ref{sec:Gaussian Rasterization and Vectorization}):
(1) a differentiable Gaussian rasterization process that renders continuous density fields for pixel-level geometric learning, and
(2) a vectorization process that directly interprets Gaussian centers as ordered vertices for topology-aware vector regression.
Finally, we describe the overall training and inference pipeline (Sec.~\ref{sec:Training and Inference}), which jointly optimizes raster and vector objectives under the same Gaussian parameterization.
Through this unified design, GSMap effectively bridges the gap between rasterized geometry and vectorized structure, achieving accurate and well-structured online HD map construction.

\subsection{Gaussian-based Representation}
\label{sec:Gaussian-based Representation}
Conventional HD map representations rely on discrete point sequences that encode only positional coordinates, inherently discarding the continuous geometric priors and spatial uncertainties present in real-world map elements. Such discretization introduces a fundamental representation gap: individual points are isolated geometric primitives that lack continuous spatial support. 

Unlike static points, Gaussians can parameterize a continuous spatial density function via center, scale, and orientation. Therefore, we re-conceptualize each map element as an ordered sequence of 2D Gaussian primitives in BEV space. This formulation allows each map element to preserve both geometric continuity and topological order within a single differentiable representation, forming the foundation for the subsequent rasterization and vectorization processes.

\noindent\textbf{Formulation.} Each 2D Gaussian $G$ is parameterized by a 5D vector:

\begin{equation}    
    G=(\mu, \sigma, \theta),
\end{equation}

where $\mu=(\mu_x,\mu_y)\in\mathbb{R}^2$ denote the center position, $\sigma=(\sigma_x, \sigma_y)\in\mathbb{R}^2$ control the scale along two principal axes, and $\theta\in\mathbb{R}$ represents the in-plane orientation. Given a spatial position $p = (x, y)$, the density of the $i$-th Gaussian is defined as:

\begin{equation}
G_i(p) = \exp\!\Big(
    -\tfrac{1}{2}\,
    (p - \mu_i)^\top
    \mathbf{\Sigma}_i^{-1}
    (p - \mu_i)
\Big),
\label{eq:gaussian_density}
\end{equation}

where the covariance matrix $\mathbf{\Sigma}_i$ is constructed as:

\begin{equation}
\mathbf{\Sigma}_i=\mathbf{R}_i\mathbf{S}_i\mathbf{S}_i^{\top}\mathbf{R}_i^{\top},
\end{equation}

with $\mathbf{S}_i = \mathrm{diag}[(\sigma_{x}^i)^2, (\sigma_{y}^i)^2]$ controlling scale, and $\mathbf{R}_i$ being the 2D rotation matrix parameterized by the learnable orientation $\theta_i$:

\begin{equation}
\mathbf{R} =
\begin{bmatrix}
\cos\theta & -\sin\theta \\
\sin\theta & \phantom{-}\cos\theta
\end{bmatrix}.
\end{equation}

For an HD map containing $M$ elements, we denote the $j$-th element as:

\begin{equation}
E^{j} = \{G_1^{j}, G_2^{j}, \dots, G_N^{j}\}, \quad j = 1, 2, \dots, M,
\end{equation}

where each map element consists of a fixed number $N$ of ordered Gaussians. 

\subsection{Gaussian Rasterization and Vectorization}
\label{sec:Gaussian Rasterization and Vectorization}
Building on the unified Gaussian-based representation, GSMap introduces two complementary pathways that connect continuous geometry with discrete topology. The rasterization pathway renders continuous BEV fields from Gaussian primitives through differentiable rendering, while the vectorization pathway reconstructs ordered geometric structures directly from Gaussian centers. Unified by a shared representation, the two pathways ensure consistent and end-to-end learnable transformation between rasterization and vectorization outputs.

\begin{figure}[t!]
    \centering
    \includegraphics[width=0.6\linewidth]{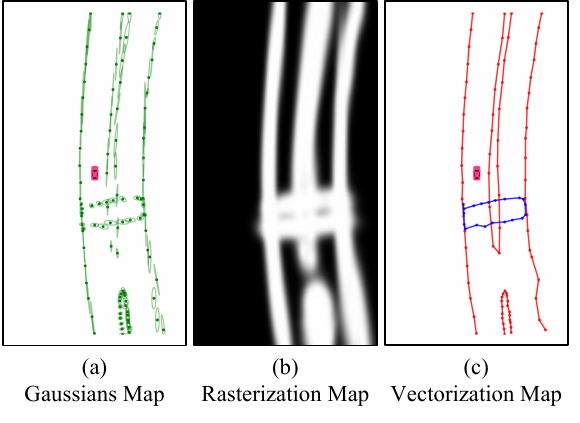}
    \caption{We propose (a) a Gaussian-based HD map representation. Two types of HD map representations are obtained through (b) rasterization and (c) vectorization. 
    (\textit{The green ellipses denote the $1\sigma$ spatial range of individual 2D Gaussians.})}
    \label{fig:Map Representation}
\end{figure}

\noindent\textbf{Rasterization.}
To obtain geometric fields at pixel-level, GSMap converts each sequence of Gaussian primitives into a continuous BEV map through a differentiable rasterization process. 
For each HD map element $E^j=\{G_1^j, G_2^j, \dots, G_N^j\}$, the rendered occupancy probability at a BEV position $p=(x,y)$ is expressed as:
\begin{equation}
\mathcal{R}_j(p) = 1 - \prod_{i=1}^{N}\big(1 - G_i^j(p)\big),
\label{eq:raster}
\end{equation}
where $G_i^j(p)$ is the Gaussian density defined in Eq.~(\ref{eq:gaussian_density}). 
This process produces a smooth and differentiable density map that continuously represents the spatial coverage of each element, removing the need for discrete polygon rasterization.

\noindent\textbf{Vectorization.}
In parallel, GSMap constructs a vectorized representation by directly using the Gaussian centers as ordered vertices. 
For each element $E^j$, the sequence of centers 
$\{\mu_i^j\}_{i=1}^{N}$ 
defines a polyline that encodes geometric continuity and topological order.
Compared with discrete point-based vectorizations that treat vertices as isolated coordinates, the Gaussian parameterization introduces local spatial support and orientation awareness through $(\sigma, \theta)$, allowing the vectorized structure to retain sub-pixel geometric precision while remaining fully differentiable.

As illustrated in Fig.~\ref{fig:Map Representation}, we visualize the Gaussian-based map representation along with its corresponding rasterized and vectorized forms. Benefiting from this unified formulation, the rasterized geometry and vectorized topology are seamlessly connected, providing a coherent foundation for joint learning within a shared representation space.

\subsection{Training and Inference}
\label{sec:Training and Inference}
\noindent\textbf{Training.}
Our GSMap is a general framework that can be seamlessly integrated with existing HD map construction models (\eg, MapTR~\cite{MapTR}). As illustrated in Fig.~\ref{fig:pipeline}, the map encoder first transforms multi-view surround images into BEV feature maps. 
During the GSMap decoder stage, we adopt a DETR-like~\cite{DETR} transformer decoder that initializes a set of instance queries for each map element.
For each instance query $q_j^{\mathrm{ins}}$, a fixed set of Gaussian queries $\{q_i^{G}\}_{i=1}^N$ is further initialized, where each query corresponds to one 2D Gaussian primitive within the element sequence.
Accordingly, the hierarchical query for the $i$-th Gaussian of the $j$-th map element is defined as:

\begin{equation}
    q_{j,i}^{H} = q_j^{\mathrm{ins}} + q_i^{G}, \quad q_{j,i}^{H} \in \mathbb{R}^{d},
\end{equation}

where $d$ denotes the feature dimension.

Following MapTR~\cite{MapTR}, we first apply a vanilla self-attention~\cite{vaswani2017attention} layer among all $\{q_{j,i}^{H}\}$, enabling each gaussian information exchange.
Subsequently, a deformable cross-attention~\cite{Deformable} layer is employed to fuse these queries with the BEV features.
The resulting queries are then passed through a lightweight feed-forward network (MLP) to predict the parameters of each 2D Gaussian:

\begin{equation}
    G_{j,i} = \mathrm{MLP}(q_{j,i}^{H}).
\end{equation}

The updated Gaussian parameters are further used to guide the feature sampling and query interaction in the next decoder layer, enabling iterative refinement of Gaussian representation.

After decoding, matching is performed at instance and point levels for the predicted Gaussian instances. At the instance level, we compute the matching cost by combining the classification score, vectorized point cost, and rasterized iou cost:
\begin{equation}
    \mathcal{C}_{\mathrm{ins\_match}} =
    \mathcal{C}_{\mathrm{cls}}
    + \mathcal{C}_{\mu}
    + \mathcal{C}_{\mathrm{iou}},
\end{equation}
where $\mathcal{C}_{\mathrm{cls}}$ and $\mathcal{C}_{\mu}$ follows MapTR~\cite{MapTR}, $\mathcal{C}_{\mathrm{iou}}$ denotes the IoU cost~\cite{MapVR} computed on rasterized Gaussian masks. 
Following the instance-level assignment, we perform point-level matching as in MapTR~\cite{MapTR} to ensure consistent alignment within each element sequence.

\noindent\textbf{Loss Functions.} After the matching process, GSMap jointly optimizes raster and vector objectives under a shared Gaussian representation. At the instance level, the supervision focuses on the correctness of category prediction and geometric fidelity of each element, formulated as:

\begin{equation}
\mathcal{L}_{\mathrm{ins}} =
\mathcal{L}_{\mathrm{cls}}
+ \lambda_{\mathrm{v}}\mathcal{L}_{\mathrm{vector}}
+ \lambda_{\mathrm{r}}\mathcal{L}_{\mathrm{raster}},
\end{equation}

where $\mathcal{L}_{\mathrm{cls}}$ is the focal loss~\cite{MapTR} for instance classification,
$\mathcal{L}_{\mathrm{vector}}$ is the vector loss~\cite{MapTR} applied to Gaussian centers,
and $\mathcal{L}_{\mathrm{raster}}$ measures the pixel-level discrepancy between rasterized Gaussian masks and ground truth masks. The weighting factors $\lambda_{\mathrm{vector}}$ and $\lambda_{\mathrm{raster}}$ balance the contributions of rasterization and vectorization supervision.

Following MapTR~\cite{MapTR}, we apply a point-to-point regression loss between predicted and ground-truth Gaussian centers. 
For each element, the vector loss is computed as:

\begin{equation}
\mathcal{L}_{\mathrm{vec}}
=
\sum_{j=1}^{M}
\sum_{i=1}^{N}
D_{\mathrm{Mht}}(
\mu_{j,i} - \mu_{j,i}^{*}
),
\end{equation}

where $D_{\mathrm{Mht}}$ denotes the Manhattan distance, 
and $\mu_{j,i}$ and $\mu_{j,i}^{*}$ are the predicted and ground-truth Gaussian centers, respectively. For rasterization supervision, each predicted element is rendered into a BEV density map $\mathcal{R}_j$ following the differentiable formulation in Eq.~\ref{eq:raster}.
The raster loss is computed as:

\begin{equation}
\mathcal{L}_{\mathrm{raster}}
=
\lambda_{\alpha}D_w(\mathcal{R}_j,\mathcal{R}_j^{*}) + (1-\lambda_{\alpha})\mathcal{L}_{\text{D-SSIM}},
\end{equation}

where $\mathcal{R}_j^{*}$ is the corresponding ground-truth mask, $D_w$ denotes a weighted $\mathcal{L}_{\text{1}}$ loss, and $\mathcal{L}_{\text{D-SSIM}}$ is the D-SSIM loss~\cite{kerbl20233d}.
This loss provides dense geometric feedback, complementing the sparse vector supervision. By jointly optimizing the rasterization and vectorization pathways, GSMap achieves consistent learning of both geometric details and structural topology.

\noindent\textbf{Inference.}
During inference, the trained model directly predicts Gaussian parameters 
${G}_{i}^{j} = ({\mu}_{i}^{j}, {\sigma}_{i}^{j}, {\theta}_{i}^{j})$ 
for each element query. 
Without any post-processing or rasterization, the ordered Gaussian centers $\{{\mu}_{i}^{j}\}_{i=1}^{N}$ are connected sequentially to form continuous vectorized map elements, such as lane dividers or road boundaries. This design enables GSMap to generate topologically consistent and geometrically precise HD maps in a single forward pass, while maintaining comparable inference efficiency to existing vectorization-based models.
\section{Experiment}
\subsection{Experiment Setup}
\textbf{Datasets.} We evaluate our method on two large-scale autonomous driving datasets: nuScenes~\cite{nuscenes} and Argoverse2~\cite{wilson2023argoverse}. The nuScenes dataset comprises 1000 driving scenes, each providing multi-view RGB images from six surround cameras. Following standard practice~\cite{nuscenes}, we split the data into 700 scenes for training and 150 for validation. Argoverse2 contains 1,000 scenes with seven-camera RGB imagery. We adopt the same split protocol as prior work~\cite{wilson2023argoverse}, using 700 scenes for training and 150 scenes for validation. Across both datasets, we focus on three map element categories that are critical for navigation: road boundaries, lane dividers, and pedestrian crossings.

\begin{table*}[ht]
\begin{center}
\centering
\caption{Comparison of various map vectorization methods on nuScenes val set.}
\label{tab:exp_result_nuscenes_basic}
\setlength{\tabcolsep}{4.25pt}
\resizebox{1.0\textwidth}{!}{
\begin{tabular}[t]{l|c|c|c|cccc|cccc|c|c}
\toprule[1.25pt]
\multirow{2}{*}{Method} & \multirow{2}{*}{Modality} & \multirow{2}{*}{Backbone} & \multirow{2}{*}{\#Epochs} & \multicolumn{4}{c|}{$\mathrm{AP}_{\mathrm{Chamfer}}\uparrow$} & \multicolumn{4}{c|}{$\mathrm{AP}_{\mathrm{raster}}\uparrow$} & \multirow{2}{*}{FPS$\uparrow$} & \multirow{2}{*}{Params} \\
\cline{5-12}
& & & & ped & div & bdry &  \textit{avg.} & ped & div & bdry &  \textit{avg.} & \\

\midrule[0.925pt]

HDMapNet~\cite{Hdmapnet} & C & Effi-B0 & 30 & 14.4 & 21.7 & 33.0 &  23.0 & - & - & - &  - & 0.7 & - \\

VectorMapNet~\cite{Vectormapnet} & C & Res-50 & 110 & 36.1 & 47.3 & 39.3 &  40.9 & 26.2 & 12.7 & 6.1 &  15.0 & 2.8 & - \\

\midrule[0.25pt]

MapTR~\cite{MapTR} & C & Res-50 & 24 &  {46.3} & 51.5 &  {53.1} &  50.3 & 32.4 & 23.5 & 17.1 &  24.3 & \textbf{17.9} & \textbf{35.92M} \\
MapTR~\cite{MapTR} + MapVR~\cite{MapVR} & C & Res-50 & 24 &  \textbf{47.7} &  {54.4} & 51.4 &   {51.2} &  {37.5} &  \textbf{33.1} &  {23.0} &  {31.2} & \textbf{17.9} & \textbf{35.92M} \\
MapTR~\cite{MapTR} + GSMap (Ours) & C & Res-50 & 24 & 45.7 &  \textbf{55.0} &  \textbf{56.2} &  \textbf{52.3} &  \textbf{39.1} &  {32.3} &  \textbf{30.2} &  \textbf{33.9} & 17.4 & 35.95M \\

\midrule[0.25pt]
MapTR~\cite{MapTR} & C & Res-50 & 110 &  \textbf{56.2} &  {59.8} &  {60.1} &  58.7 & 43.6 & 35.6 & 25.8 &  35.0 & \textbf{17.9} & \textbf{35.92M}\\
MapTR~\cite{MapTR} + MapVR~\cite{MapVR} & C & Res-50 & 110 & 55.0 &  \textbf{61.8} & 59.4 &  {58.8} &  {46.0} &  {39.7} &  {29.9} &  {38.5} & \textbf{17.9} & \textbf{35.92M} \\
MapTR~\cite{MapTR} + GSMap (Ours) & C & Res-50 & 110 &  {55.8} &  \textbf{61.8} &  \textbf{61.9} &  \textbf{59.8} &  \textbf{50.7} &  \textbf{40.5} &  \textbf{35.9} &  \textbf{42.4} & 17.4 & 35.95M \\
\bottomrule[1.25pt]
\end{tabular}}
\end{center}
\end{table*}

\noindent\textbf{Evaluation Metric.}
For fair comparison, we adopt two commonly used metrics: the Chamfer distance-based average precision $\mathrm{AP}_{\mathrm{chamfer}}$ from MapTR~\cite{MapTR}, and the IoU-based average precision $\mathrm{AP}_{\mathrm{raster}}$ from MapVR~\cite{MapVR}. 
Specifically, $\mathrm{AP}_{\mathrm{chamfer}}$ is computed as the average precision averaged over three Chamfer distance thresholds $\tau_{\mathrm{chamfer}}\in\{0.5\text{m}, 1.0\text{m}, 1.5\text{m}\}$. 
Following MapVR, $\mathrm{AP}_{\mathrm{raster}}$ is evaluated with IoU thresholds of $\{0.25\!:\!0.5\!:\!0.05\}$ for road boundaries and lane dividers, and $\{0.5\!:\!0.75\!:\!0.05\}$ for pedestrian crossings.

\begin{table}[h]
\centering
\caption{Comparison of various map vectorization methods on Argoverse2 val set.}
\footnotesize
\setlength{\tabcolsep}{3pt}
\renewcommand{\arraystretch}{1.05}

\newcolumntype{Y}{>{\centering\arraybackslash}X} 

\begin{tabularx}{0.8\columnwidth}{l|YYYY}
\toprule[1.25pt]
\multirow{2}{*}{Method} & \multicolumn{4}{c}{$\mathrm{AP}_{\mathrm{Chamfer}}\uparrow$} \\
\cline{2-5}
& ped & div & bdry & \textit{avg.} \\
\midrule[0.925pt]
HDMapNet~\cite{Hdmapnet} & 13.1 & 5.7 & 37.6 & 18.8 \\
VectorMapNet~\cite{Vectormapnet} & 38.3 & 36.1 & 39.2 & 37.9 \\
MapTR~\cite{MapTR} & 54.7 & 58.1 & 56.7 & 56.5 \\
MapTR\,\cite{MapTR} + MapVR & 54.6 & 60.0 & 58.0 & 57.5 \\
MapTR\,\cite{MapTR} + GSMap\,(Ours) & \textbf{57.0} & \textbf{60.2} & \textbf{60.6} & \textbf{59.2} \\
\bottomrule[1.25pt]
\end{tabularx}
\vspace{10pt}
\label{tab:exp_result_argoverse2}
\end{table}

\noindent\textbf{Implementation Details.} 
All experiments are conducted on 8 NVIDIA GeForce RTX 3090 GPUs with a total batch size of 2. 
The initial learning rate is set to $6\times10^{-4}$ and decayed following a cosine annealing schedule. 
The BEV feature map is configured to $200\times100$ resolution, corresponding to a perception range of $[-30\text{m}, 30\text{m}]$ longitudinally and $[-15\text{m}, 15\text{m}]$ laterally. 
We employ $M=50$ Gaussian instance queries to detect map elements, with each instance represented by $N=20$ 2D Gaussian primitives. The model is trained on the nuScenes dataset for 24 and 110 epochs to ensure fair comparison with previous methods, and on the Argoverse 2 dataset for 6 epochs for validation.

\subsection{Experiment Results}
\noindent\textbf{Results on nuScenes.}
Tab.~\ref{tab:exp_result_nuscenes_basic} summarizes our comparison on the nuScenes Map validation split.
Our framework is evaluated by integrating GSMap into the MapTR baseline under two training schedules (24 and 110 epochs).
At both settings, GSMap consistently improves performance across all categories and both evaluation metrics.
Specifically, under the 24-epoch schedule, GSMap achieves an average $\mathrm{AP}_{\mathrm{Chamfer}}$ of 52.3 and $\mathrm{AP}_{\mathrm{raster}}$ of 33.9, surpassing MapTR by 2.0 and 9.6, and outperforming MapTR+MapVR by 1.1 and 2.7, respectively. With a 110-epoch schedule, GSMap further improves to 59.8 and 42.4, exceeding MapTR by 1.1 and 7.4, and MapTR+MapVR by 1.0 and 3.9, respectively.

Notably, these improvements are achieved with nearly identical inference speed and parameter size to the MapTR baseline, showing that GSMap enhances geometric and topological learning without introducing additional computational overhead. The consistent improvements in both Chamfer-based and IoU-based metrics confirm that the proposed Gaussian representation effectively unifies rasterized geometry and vectorized structure within a single, learnable formulation. Moreover, compared with MapVR, which uses rasterization merely as auxiliary supervision, GSMap more effectively jointly optimizes geometric accuracy and structural consistency.

\begin{figure*}[ht]
    \centering
    \includegraphics [width=1\textwidth]{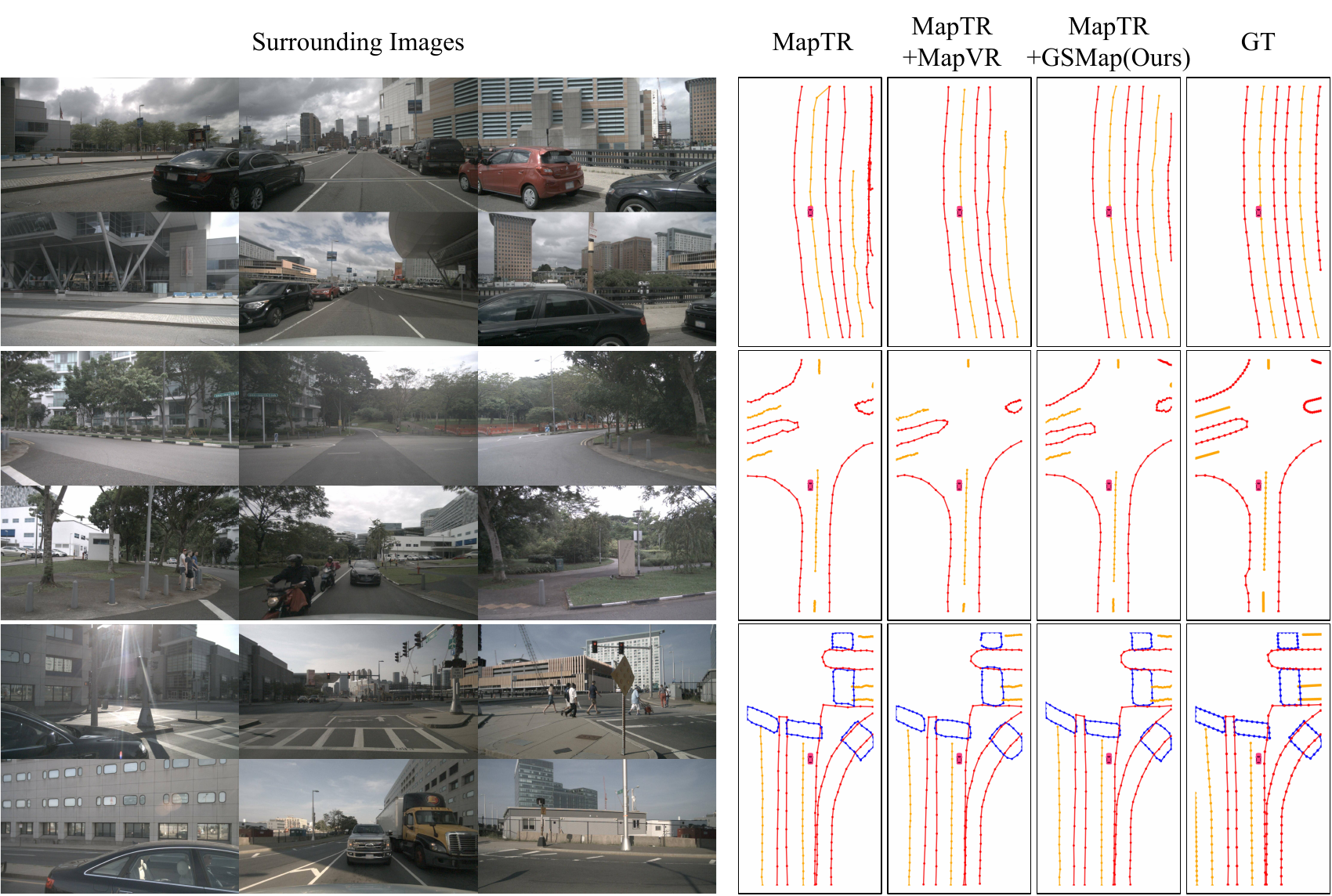}
    \caption{Visualization of online HD map vectorization results on nuScenes val set.}
    \label{fig:vis}
\end{figure*}

\noindent\textbf{Results on Argoverse 2.}
Tab.~\ref{tab:exp_result_argoverse2} presents the comparison results on the Argoverse 2 validation set.
Similar to the observations on nuScenes, integrating GSMap into MapTR yields consistent and notable improvements across all categories.
GSMap achieves the highest overall performance, reaching an average $\mathrm{AP}_{\mathrm{Chamfer}}$ of 59.2, outperforming MapTR by 2.7 and MapTR+MapVR by 1.7.

\begin{table}[h]
\centering
\caption{Ablation on different baselines on the nuScenes dataset. Integrating GSMap consistently improves $\mathrm{AP}_{\mathrm{Chamfer}}$. $^\dagger$ denotes that 50 one-to-many queries are used for a fair comparison under identical GPU memory settings.}
\footnotesize
\setlength{\tabcolsep}{3pt}
\renewcommand{\arraystretch}{1.05}

\newcolumntype{Y}{>{\centering\arraybackslash}X} 

\begin{tabularx}{0.8\columnwidth}{l|YYYY}
\toprule[1.25pt]
\multirow{2}{*}{Method} & \multicolumn{4}{c}{$\mathrm{AP}_{\mathrm{Chamfer}}\uparrow$} \\
\cline{2-5}
& ped & div & bdry & \textit{avg.} \\
\midrule[0.925pt]
MapTR~\cite{MapTR} & 46.3 & 51.5 & 53.1 & 50.3 \\
\cdashline{1-5}[2.5pt/5pt]
MapTR~\cite{MapTR} + GSMap(Ours) & \makecell{45.7 \\ \textcolor{red!80!black}{\scriptsize (-0.6)}} & \makecell{55.0 \\ \textcolor{green!50!black}{\scriptsize (+3.5)}} & \makecell{56.2 \\ \textcolor{green!50!black}{\scriptsize (+3.1)}} & \makecell{52.3 \\ \textcolor{green!50!black}{\scriptsize (+2.0)}} \\
\midrule[0.25pt]
MapTRv2$^\dagger$~\cite{Maptrv2} & 57.2 & 59.8 & 62.0 & 59.6 \\
\cdashline{1-5}[2.5pt/5pt]
MapTRv2$^\dagger$~\cite{Maptrv2} + GSMap(Ours) & \makecell{58.6 \\ \textcolor{green!50!black}{\scriptsize (+1.4)}} & \makecell{61.1 \\ \textcolor{green!50!black}{\scriptsize (+1.3)}} & \makecell{63.6 \\ \textcolor{green!50!black}{\scriptsize (+1.6)}} & \makecell{61.1 \\ \textcolor{green!50!black}{\scriptsize (+1.5)}} \\
\bottomrule[1.25pt]
\end{tabularx}
\vspace{10pt}
\label{tab:ablation_baseline}
\end{table}

\subsection{Ablation study}
All ablation experiments are conducted on the MapTR baseline trained for 24 epochs to ensure a fair and efficient comparison.  
We systematically analyze the impact of different integration strategies, loss formulations, hyperparameters, and architectural configurations to validate the effectiveness and flexibility of the proposed GSMap framework.

\noindent\textbf{Ablation on Baseline.}  
This experiment evaluates the effect of integrating GSMap into different HD map baselines, including MapTR and MapTRv2. As shown in Tab.~\ref{tab:ablation_baseline}, GSMap consistently improves performance across both baselines in terms of $\mathrm{AP}_{\mathrm{Chamfer}}$. Specifically, integrating GSMap into MapTR increases the average $\mathrm{AP}_{\mathrm{Chamfer}}$ by 2.0, with notable improvements of 3.5 on the divider category and 3.1 on the boundary category. When integrated with MapTRv2, GSMap further raises the overall average by 1.5, improving all categories consistently. These results demonstrate that GSMap can be seamlessly incorporated into various HD map frameworks, providing stable and generalizable performance improvements through its unified Gaussian-based representation.

\begin{table}[ht]
\centering
\caption{Ablation on the effect of rasterization loss on the nuScenes dataset.}
\footnotesize
\setlength{\tabcolsep}{3pt}
\renewcommand{\arraystretch}{1.05}

\newcolumntype{Y}{>{\centering\arraybackslash}X} 

\begin{tabularx}{0.8\columnwidth}{c|c|YYYY|YYYY}
\toprule[1.25pt]
\multirow{2}{*}{$\mathcal{L}_{\text{vector}}$} &
\multirow{2}{*}{$\mathcal{L}_{\text{raster}}$} &
\multicolumn{4}{c|}{$\mathrm{AP}_{\mathrm{Chamfer}}\uparrow$} &
\multicolumn{4}{c}{$\mathrm{AP}_{\mathrm{raster}}\uparrow$} \\
\cline{3-10}
& & ped & div & bdry & \textit{avg.} & ped & div & bdry & \textit{avg.} \\
\midrule[0.925pt]
\checkmark &  & 42.6 & 53.9 & 52.6 & 49.7 & 36.0 & 30.8 & 28.6 & 31.8 \\
\checkmark & \checkmark & \textbf{45.7} & \textbf{55.0} & \textbf{56.2} & \textbf{52.3} &
\textbf{39.1} & \textbf{32.3} & \textbf{30.2} & \textbf{33.9} \\
\bottomrule[1.25pt]
\end{tabularx}
\vspace{10pt}
\label{tab:ablation_losses}
\end{table}

\begin{figure}[ht]
    \centering
    \includegraphics [width=0.62\linewidth]{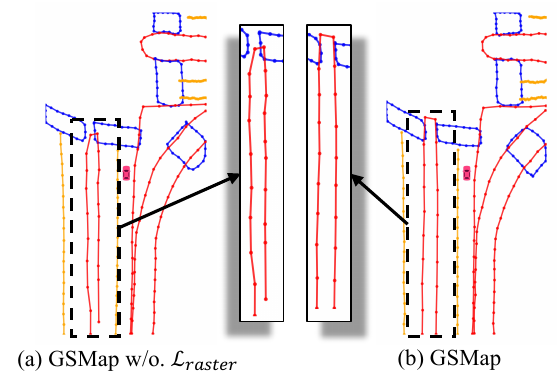}
    \caption{Effect of rasterization loss on HD map predictions. (a) GSMap without $\mathcal{L}_{\text{raster}}$ produces distorted and less accurate boundaries. (b) GSMap generates smoother and more faithful boundaries, highlighting the contribution of raster-level supervision in refining geometric fidelity and topological consistency.}
    \label{fig:raster_ablation}
\end{figure}

\noindent\textbf{Ablation of Loss Design.}  
We evaluate the contribution of the rasterization loss $\mathcal{L}_{\text{raster}}$ alongside the vectorization-based loss $\mathcal{L}_{\text{vector}}$.  
As reported in Tab.~\ref{tab:ablation_losses}, removing $\mathcal{L}_{\text{raster}}$ causes a clear performance decline across all map element types, with the most pronounced decrease observed for road boundaries, where $\mathrm{AP}_{\mathrm{Chamfer}}$ drops from 56.2 to 52.6 and $\mathrm{AP}_{\mathrm{raster}}$ from 30.2 to 28.6. Pedestrian crossings and lane dividers are also affected. This demonstrates that rasterization supervision is particularly important for capturing fine-grained geometric details of elongated or irregular structures. The effect is illustrated in Fig.~\ref{fig:raster_ablation}, where predictions without $\mathcal{L}_{\text{raster}}$ exhibit jagged boundaries, while including the rasterization loss produces smoother and more accurate boundaries, confirming that rasterization supervision enhances both geometric fidelity and topological consistency.

\begin{table}[ht]
\centering
\caption{Ablation on the loss weight $\lambda_r$ balancing rasterization and vectorization losses on the nuScenes dataset.}
\footnotesize
\setlength{\tabcolsep}{3pt}
\renewcommand{\arraystretch}{1.05}

\newcolumntype{Y}{>{\centering\arraybackslash}X}

\begin{tabularx}{0.8\columnwidth}{Y|YYYY|YYYY}
\toprule[1.25pt]
\multirow{2}{*}{$\lambda_r$} &
\multicolumn{4}{c|}{$\mathrm{AP}_{\mathrm{Chamfer}}\uparrow$} &
\multicolumn{4}{c}{$\mathrm{AP}_{\mathrm{raster}}\uparrow$} \\
\cline{2-9}
& ped & div & bdry & \textit{avg.} & ped & div & bdry & \textit{avg.} \\
\midrule[0.925pt]
5 & 40.3 & 52.8 & 54.1 & 49.1 & 37.2 & 31.3 & 28.3 & 32.3 \\
10 & \textbf{45.7} & \textbf{55.0} & \textbf{56.2} & \textbf{52.3} & \textbf{39.1} & \textbf{32.3} & \textbf{30.2} & \textbf{33.9} \\
15 & 44.3 & 51.8 & 55.4 & 50.5 & 38.5 & 30.8 & 28.8 & 32.7 \\
\bottomrule[1.25pt]
\end{tabularx}
\vspace{10pt}
\label{tab:ablation_weight}
\end{table}
\noindent\textbf{Ablation of Loss Weight.}  
We fix the vectorization loss weights $\lambda_v$ and vary the weighting coefficient $\lambda_r$ to balance the vectorization and rasterization supervision.
As shown in Tab.~\ref{tab:ablation_weight}, $\lambda_r{=}10$ yields the best overall performance ($\mathrm{AP}_{\mathrm{Chamfer}}{=}$ 52.3, $\mathrm{AP}_{\mathrm{raster}}{=}$ 33.9).
A smaller weight $\lambda_r{=}5$ weakens raster guidance, while a larger one $\lambda_r{=}15$ overemphasizes it, both leading to slight performance drops.
These results suggest that moderate rasterization supervision effectively regularizes training without affecting the vectorization objectives.

\begin{table}[ht]
\centering
\caption{Ablation on the number of Gaussian primitives per map element on the nuScenes dataset.}
\footnotesize
\setlength{\tabcolsep}{3pt}
\renewcommand{\arraystretch}{1.05}

\newcolumntype{Y}{>{\centering\arraybackslash}X} 

\begin{tabularx}{0.8\columnwidth}{c|YYYY|YYYY}
\toprule[1.25pt]
\multirow{2}{*}{$Num_{\text{Gaussian}}$} &
\multicolumn{4}{c|}{$\mathrm{AP}_{\mathrm{Chamfer}}\uparrow$} &
\multicolumn{4}{c}{$\mathrm{AP}_{\mathrm{raster}}\uparrow$} \\
\cline{2-9}
& ped & div & bdry & \textit{avg.} & ped & div & bdry & \textit{avg.} \\
\midrule[0.925pt]
10 & 40.9 & \textbf{55.1} & 53.4 & 49.8 & 37.1 & 32.3 & 27.9 & 32.4 \\
20 & \textbf{45.7} & 55.0 & \textbf{56.2} & \textbf{52.3} & \textbf{39.1} & \textbf{32.3} & \textbf{30.2} & \textbf{33.9} \\
30 & 42.7 & 51.8 & 54.0 & 49.5 & 35.8 & 29.7 & 29.2 & 31.6 \\
\bottomrule[1.25pt]
\end{tabularx}
\vspace{10pt}
\label{tab:ablation_gsnum}
\end{table}

\noindent\textbf{Ablation of Gaussian Quantity.}  
We study the effect of the number of Gaussian primitives per map element. Using too few Gaussians limits the ability to represent complex geometry, while too many provides little additional benefit. A configuration of 20 Gaussians per element achieves strong performance, with an average $\mathrm{AP}_{\mathrm{Chamfer}}$ of 52.3 and $\mathrm{AP}_{\mathrm{raster}}$ of 33.9.

\begin{table}[ht]
\centering
\caption{Ablation on rasterization resolution on the nuScenes dataset.}
\footnotesize
\setlength{\tabcolsep}{3pt}
\renewcommand{\arraystretch}{1.05}

\newcolumntype{Y}{>{\centering\arraybackslash}X} 

\begin{tabularx}{0.9\columnwidth}{c|YYYY|YYYY|c}
\toprule[1.25pt]
\multirow{2}{*}{Resolution} &
\multicolumn{4}{c|}{$\mathrm{AP}_{\mathrm{Chamfer}}\uparrow$} &
\multicolumn{4}{c|}{$\mathrm{AP}_{\mathrm{raster}}\uparrow$} &
\multirow{2}{*}{GPU Memory} \\
\cline{2-9}
& ped & div & bdry & \textit{avg.} & ped & div & bdry & \textit{avg.} \\
\midrule[0.925pt]
128 $\times$ 64  & 41.4 & 53.5 & 55.2 & 50.0 & 38.0 & 31.5 & 29.0 & 32.8 & 8,012MB \\
200 $\times$ 100 & \textbf{45.7} & \textbf{55.0} & \textbf{56.2} & \textbf{52.3} &
          \textbf{39.1} & \textbf{32.3} & \textbf{30.2} & \textbf{33.9} & 11,615MB \\
256 $\times$ 128 & 44.8 & 54.4 & 54.8 & 51.3 & 38.5 & 31.6 & 28.8 & 33.0 & 15,987MB \\
\bottomrule[1.25pt]
\end{tabularx}
\vspace{10pt}
\label{tab:ablation_rasterization_resolution}
\end{table}

\noindent\textbf{Ablation of Rasterization Resolution.}  
Finally, we evaluate how the BEV rasterization resolution influences learning.  
Tab.~\ref{tab:ablation_rasterization_resolution} indicates that overly coarse resolutions lose structural details, while excessively fine ones provide limited additional benefit but increase computational cost.  
The configuration of $200 \times 100$ achieves the best balance between accuracy and efficiency.

\section{Conclusion}
We introduce GSMap, a unified Gaussian-based representation for online HD map construction. Distinct from previous approaches that separately handle vectorized structure and rasterized geometry, GSMap represents each map element as a set of learnable 2D Gaussians and jointly optimizes both via a differentiable rasterization loss. This paradigm bridges geometric fidelity and topological consistency, capturing continuous, fine-grained map details. Experiments on nuScenes and Argoverse 2 show that GSMap consistently outperforms the base model across all categories under Chamfer and IoU metrics, while maintaining similar inference speed and parameter size. It can be integrated with various HD map frameworks such as MapTR and MapTRv2, offering a general, extensible solution for high-definition map learning.

\bibliographystyle{splncs04}
\bibliography{main}
\end{document}